\documentclass{article}
\usepackage{spconf,amsmath,graphicx,hyperref}
\usepackage{lipsum}
\usepackage{amssymb}
\usepackage{booktabs}
\usepackage{xcolor}
\usepackage[table]{xcolor}
\usepackage{makecell}
\usepackage{multirow}
\usepackage{graphicx}

\title{EgoPressDiff: Multimodal Video Diffusion for Egocentric UV-Domain Hand-Pressure Estimation}

\name{Yuan Zeng\textsuperscript{1} \qquad Zilue Gao\textsuperscript{1} \qquad Yujia Shi\textsuperscript{2,3} \qquad Zongqing Lu\textsuperscript{1} \qquad Wenming Yang\textsuperscript{1} \qquad QingMin Liao\textsuperscript{1}$^{\dagger}$
\thanks{ $\dagger$ Corresponding author.}}
\address{\textsuperscript{1}Shenzhen International Graduate School, Tsinghua University, China\\
\textsuperscript{2}Harbin Institute of Technology, China\\
\textsuperscript{3}Pengcheng Laboratory, China}

\begin{document}
\ninept
\maketitle
\begin{abstract}
Estimating hand-surface contact pressure from an egocentric view is crucial for AR/VR devices, robotic imitation, and ergonomic analysis. Existing methods often discretize pressure signal and process frames independently, leading to quantization errors and temporal inconsistencies.
We present \emph{EgoPressDiff}, a conditional video diffusion framework that generates UV-pressure maps from visual input.
The core of our approach is a multi-modal conditioning strategy, introducing a PoseNet and a Vertex Encoder to efficiently extract features from hand pose and 3D mesh vertices. These signals, along with depth information, guide the generative process to ensure the pressure fields are physically grounded. To effectively fuse these heterogeneous features, we further propose a Distribution-Calibrated Spatial Layer, which aligns their statistical properties before combination.
Evaluated on the EgoPressure ego-view setting, EgoPressDiff achieves state-of-the-art results, improving Volumetric IoU by over 34\% relative to prior baseline, while reducing MAE and maintaining high temporal accuracy. Our project page is at 
\href{https://egopressdiff.github.io/}{https://EgoPressDiff.github.io/}

\end{abstract}


\begin{figure}[h] \centering
    \includegraphics[width=0.45\textwidth]{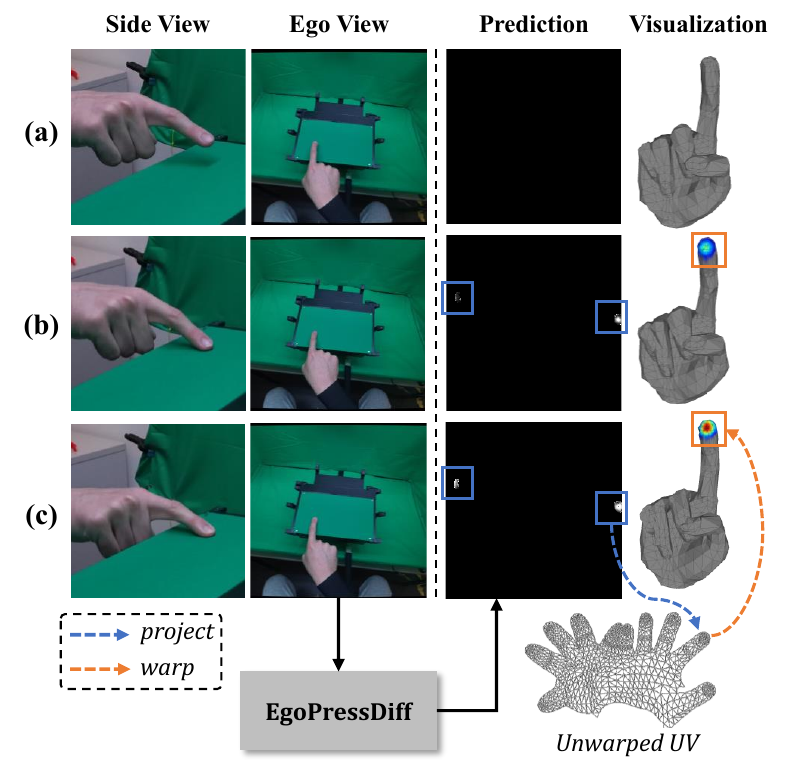}
    \caption{
    EgoPressDiff generates dynamic hand pressure from a single egocentric RGB video. It outputs a sequence of UV-pressure maps that could be warped onto the 3D MANO model for intuitive 3D visualization of contact forces. The ``Side View" is provided to better illustrate the hand’s contact with the touchpad.
    }
    \label{fig:figure1}
\end{figure}

\section{Introduction}
\label{sec:intro}
Estimating hand-surface contact pressure from an egocentric camera is a core challenge in understanding human-object interaction \cite{Ego4d, EgoExo4d}. Such information is vital for a range of applications: it provides rich, nuanced input for Augmented Reality \cite{PlayAnywhere, Opportunistic} / Virtual Reality \cite{MRTouch} systems, assists in robotic imitation \cite{Dgrasp, collins2023visual}, and supports detailed ergonomic assessments. While direct measurement requires cumbersome sensors \cite{luo2021learning, zlokapa2022integrated}, estimating dense pressure from vision offers a scalable, non-intrusive alternative. 

Recent work shows promising progress with encoder-decoder pipelines. 
Seminal works like PressureVision \cite{PressureVision} demonstrated the feasibility of predicting 2D pressure maps from a single RGB image by classifying pixels into discrete pressure bins. 
In EgoPressure \cite{Egopressure}, an extension termed PressureVision with HaMeR \cite{hamer} augments PressureVision with a 2.5D hand-pose channel to improve performance. 
Furthermore, PressureVision++ \cite{Pressurevision++} improves generalization by using a multi-task framework with weak ``contact labels" and adversarial domain adaptation.
Recently, PressureFormer \cite{Egopressure} presents a different approach by using a transformer-based decoder to predict pressure directly as a UV map on the 3D hand mesh, rather than in the 2D image plane.
Despite these advances, existing methods share three critical limitations:
(i) They discretize continuous pressure into a small number of bins and train a per-pixel classifier. This introduces quantization errors and fails to capture subtle pressure changes.
(ii) They operate on individual frames, ignoring the crucial temporal context of an action (e.g., press, peak, and release). This leads to temporally inconsistent predictions.
(iii) They rely primarily on RGB data with limited geometric priors, failing to fully exploit structured conditions like depth, skeletal kinematics, and 3D mesh geometry. This constrains their ability to infer the underlying physics of contact from visual cues alone.

\begin{figure*}[tb] \centering
    \includegraphics[width=0.93\textwidth]{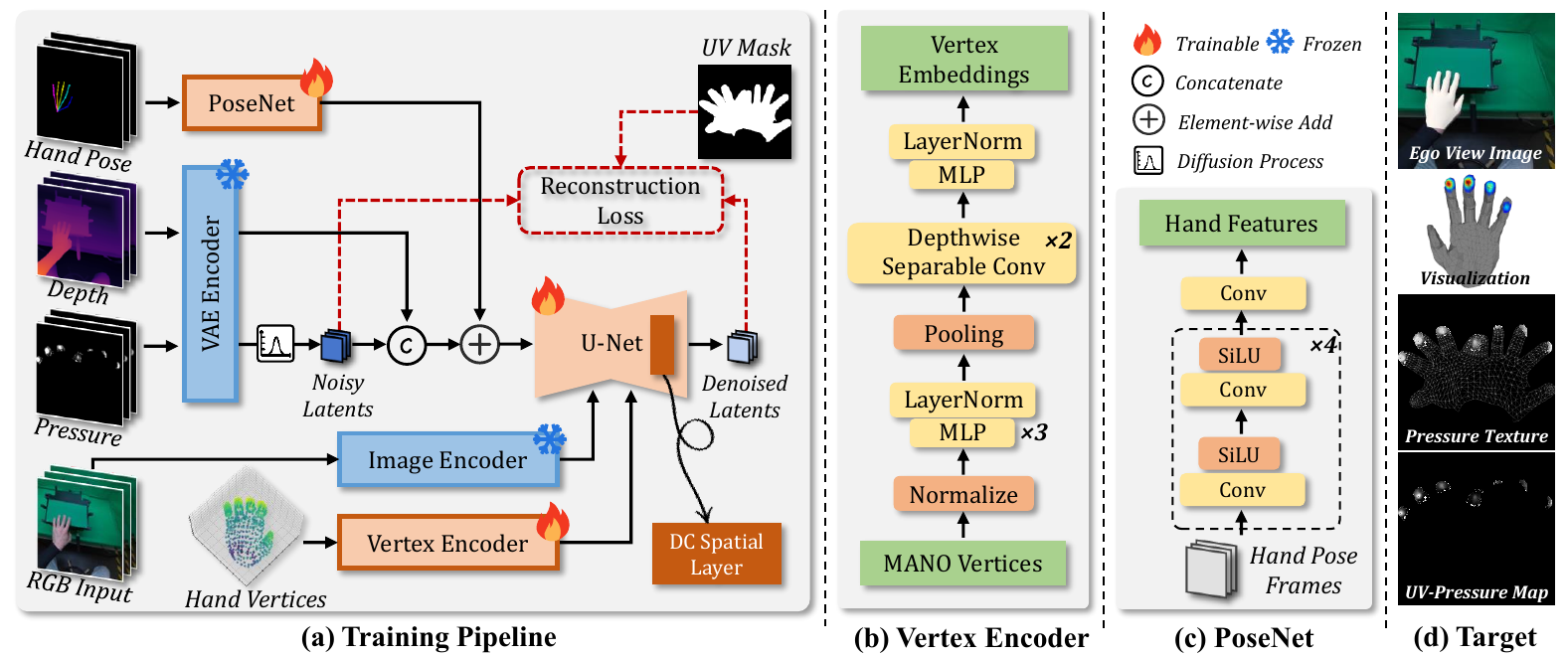}
    \caption{
    Overview of EgoPressDiff.
    \textbf{(a)} The training pipeline of EgoPressDiff. 
    The model processes five input streams through dedicated encoders: a PoseNet for hand pose, a VAE for depth and UV-pressure, a CLIP image encoder for RGB frames, and a Vertex Encoder for hand vertices. To align features from the vertex and image embeddings, we introduce a DC Spatial Layer, which replaces the original spatial layer in the U-Net. The pipeline is trained end-to-end using a reconstruction loss, with a UV mask employed to up-weight the hand UV pixels.
    \textbf{(b)} The architecture of Vertex Encoder.
    \textbf{(c)} The architecture of PoseNet.
    \textbf{(d)} Visualization of the reconstruction target, from top to bottom: The input egocentric RGB image with 3D MANO hand mesh; A 3D visualization of the ground-truth pressure on the hand surface; The pressure represented as a texture on the unwrapped UV layout; The final 2D UV-pressure map, which serves as the reconstruction target. 
    }
    \label{fig:figure2}
\end{figure*}

To address these challenges, we reframe the problem from static, per-pixel classification to continuous and dynamic video generation. Building on this formulation, we introduce EgoPressDiff, a conditional video diffusion framework \cite{svd} designed to generate UV-pressure maps from egocentric video. As shown in Figure~\ref{fig:figure1}, EgoPressDiff outputs a temporally coherent sequence of UV-pressure maps that could be textured onto a MANO \cite{mano} mesh for intuitive 3D visualization.
Our method centers on a multi-modal conditioning strategy. The generative process is guided by diverse control signals, including hand pose maps, depth maps, MANO \cite{mano} vertex sequences, and RGB frames, ensuring the generated pressure fields are both visually plausible and physically grounded.
By modeling pressure as a continuous spatiotemporal process with video diffusion models, EgoPressDiff overcomes the limitations of prior work, producing smoother, temporally stable, and physically consistent results.

Our main contributions are summarized as follows:
(1) To our knowledge, we are the first to apply a video diffusion model to the task of hand pressure estimation.
(2) Our conditioning mechanism grounds pressure generation in physical kinematics by fusing multi-modal geometric priors (depth, pose, and 3D vertices) with visual features.
(3) Our method sets a new state-of-the-art on the EgoPressure ego-view setting, demonstrating the superiority of our continuous and temporally-aware modeling.


\section{Methods}
\label{sec:Methods}


\subsection{Network Architecture}

\textbf{Overview.}
In this section, we elaborate the architecture of our model. 
The training pipeline of our method is illustrated in Figure~\ref{fig:figure2} (a). The network consists of several key components, including the PoseNet, Vertex Encoder, and Distribution-Calibrated (DC) Spatial Layer. These modules work together to extract, fuse, and align multi-modal features. First, PoseNet is employed to extract hand pose features, which are added to the input latent to explicitly enhance the model's awareness of the hand's posture. Second, since depth information is critical for inferring hand contact with the environment, we use VAE to encode the depth maps, projecting its features into the same feature space as input latents. To further capture the hand's spatial structure, we propose a Vertex Encoder that uses the 3D vertex coordinates of the MANO hand model \cite{mano} as a geometric prior, enabling the model to learn the correlation between 3D hand geometry and pressure. Furthermore, a CLIP \cite{clip} image encoder processes the egocentric RGB frames. The resulting image embeddings serve a dual purpose: they provide global visual context and, within the DC Spatial Layer, calibrate the distribution of hand-vertex embeddings, ensuring distributional consistency across modalities.
In what follows, we will present the architectural details of these key components.

\noindent \textbf{Vertex Encoder.}
To efficiently extract a compact and temporally coherent representation from the MANO hand vertices, we propose a Vertex Encoder. 
The input is a sequence of hand-mesh vertices with shape $(B, N, 778, 3)$, where $B$, $N$, $778$, and $3$ correspond to the batch size, number of frames, number of MANO vertices, and $(x,y,z)$ coordinates, respectively. The encoder outputs a feature sequence of shape $(B, N, 1024)$. 
As illustrated in Figure~\ref{fig:figure2} (b), the encoding process consists of three main stages. First, for spatial feature extraction, the input vertices for each frame undergo a normalization step to remove global translation and scale, enhancing invariance. Subsequently, a shared MLP block is applied to each of the 778 vertices independently, mapping their 3D coordinates to a higher-dimensional feature space. To aggregate these point-wise features into a single, permutation-invariant frame-level embedding, we apply a symmetric max-pooling operation across the vertex dimension. Second, for temporal feature mixing, the resulting embeddings is processed by a lightweight temporal module. This module consists of stacked depthwise-separable convolutional blocks, which effectively capture local temporal dynamics with minimal computational overhead. Finally, a linear projection layer followed by a LayerNorm \cite{layerNorm} maps the temporally-aware features to the final 1024-dimensional output. This hierarchical design allows the Vertex Encoder to efficiently distill the MANO vertex data into a rich, structured representation suitable for denoising models.

\noindent \textbf{PoseNet.}
Although many generative models integrate human pose features into denoising models via ControlNet \cite{ControlNet}, this strategy incurs substantial computational overhead. Inspired by AnimateAnyone \cite{Animateanyone}, we introduce a lightweight PoseNet for extracting hand pose map features. As depicted in Figure~\ref{fig:figure2} (c), our PoseNet consists solely of convolutional and SiLU \cite{silu} layers. To ensure stable training, the network is initialized with Gaussian weights, and its final projection layer is implemented as a zero-initialized convolution. 

\begin{figure}[h] \centering
    \includegraphics[width=0.48\textwidth]{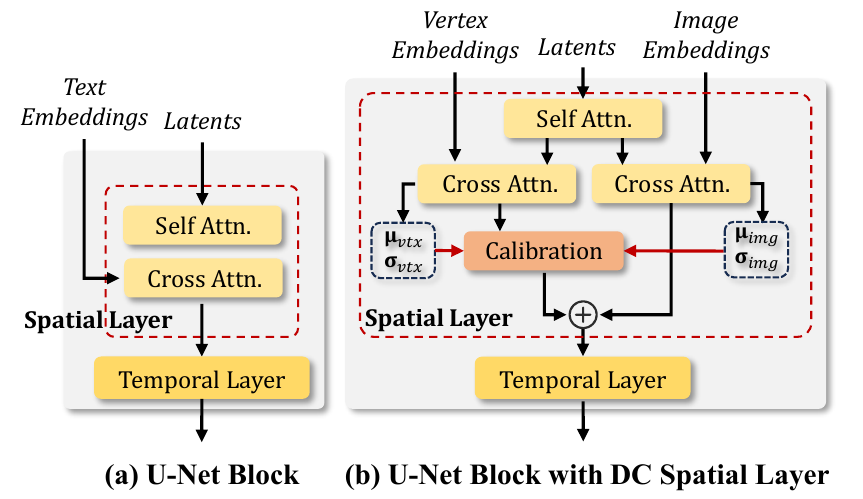}
    \caption{(a) The original U-Net block. (b) Our proposed Distribution-Calibrated (DC) Spatial Layer integrated into the U-Net block. Here, $\mu$ and $\sigma$ denote the mean and standard deviation, respectively.}
    \label{fig:figure3}
\end{figure}

\noindent \textbf{Distribution-Calibrated Spatial Layer.}
To effectively integrate multimodal control signals, we introduce the DC Spatial Layer. As shown in Figure~\ref{fig:figure3} (a), the spatial layer in a standard diffusion U-Net \cite{UNet} conditions latent features on text embeddings via a single cross-attention block. In contrast, our DC Spatial Layer, depicted in Figure~\ref{fig:figure3} (b), replaces this text branch with a dual-branch design that simultaneously conditions on image and vertex embeddings. However, because these embeddings originate from different domains, the latents produced by their respective cross-attention blocks tend to occupy disparate feature spaces. We therefore introduce a calibration block that aligns the two latent distributions prior to fusion, enabling stable and effective cross-modal conditioning.

Let $\boldsymbol{z}$ be the latent input to the spatial layer. After a self-attention block, $\boldsymbol{z}$ is fed into two parallel cross-attention blocks, conditioned on the image and vertex embeddings, respectively. This step yields two intermediate latents, $\boldsymbol{z}^{img}$ and $\boldsymbol{z}^{vtx}$, which exhibit different statistical distributions.
In calibration block, we compute the channel-wise mean and standard deviation for both latents, denoted as $(\boldsymbol{\mu}_{img}, \boldsymbol{\sigma}_{img})$ and $(\boldsymbol{\mu}_{vtx}, \boldsymbol{\sigma}_{vtx})$. We then align the latents by enforcing the equivalence of their standardized representations:
\begin{equation}
\frac{\boldsymbol{z}^{img} - \boldsymbol{\mu}_{img}}{\boldsymbol{\sigma}_{img}} = \frac{\boldsymbol{z}^{vtx} - \boldsymbol{\mu}_{vtx}}{\boldsymbol{\sigma}_{vtx}}.
\end{equation}
From this, we derive the calibrated vertex latent $\bar{\boldsymbol{z}}^{vtx}$ by remapping $\boldsymbol{z}^{vtx}$ to match the statistical properties of $\boldsymbol{z}^{img}$:
\begin{equation}
\bar{\boldsymbol{z}}^{vtx} = \frac{\boldsymbol{z}^{vtx} - \boldsymbol{\mu}_{vtx}}{\boldsymbol{\sigma}_{vtx}} \times \boldsymbol{\sigma}_{img} + \boldsymbol{\mu}_{img}.
\end{equation}
Finally, the calibrated $\bar{\boldsymbol{z}}^{vtx}$ is fused with $\boldsymbol{z}^{img}$ via element-wise addition, and the result is passed to the subsequent temporal layer. 

\subsection{Training Loss}
As illustrated in Figure~\ref{fig:figure2} (a), the model is trained end-to-end with a reconstruction loss. The loss is optimized over the trainable parameters of the UNet, Vertex Encoder, and PoseNet. To enhance the physical plausibility of the output, we introduce UV mask as a spatial prior in the loss computation. This mask compels the model to prioritize the reconstruction accuracy within the valid hand region defined by the UV map. The training objective is formulated as a weighted mean squared error:
\begin{equation}
\mathcal{L} = \mathbb{E}_{\varepsilon}\left(\left\|\left(z_{gt}-z_{\varepsilon}\right)\odot\left(1+\mathbf{M}_{uv}\right)\right\|^2\right),
\end{equation}
where $\boldsymbol{z}_{gt}$ is the ground truth latent, $\boldsymbol{z}_{\varepsilon}$ is the denoised latent, and $\mathbf{M}_{uv}$ is the binary UV mask. 

\begin{figure*}[tb] \centering
    \includegraphics[width=0.95\textwidth]{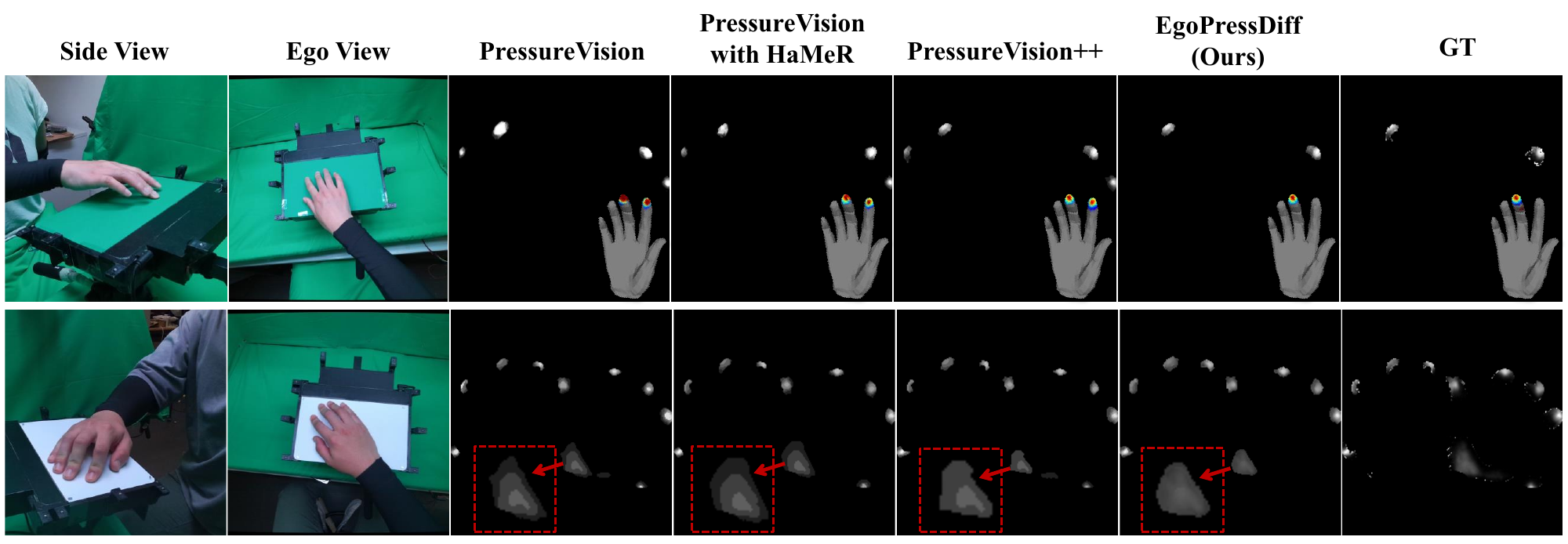}
    \caption{
    Qualitative comparison on the EgoPressure dataset.
    The ``Ego View" serves as the input, while the ``Side View" is provided to better illustrate the hand's contact with the touchpad. The first row shows a case where only the middle finger is in contact and only our method predicts this correctly. In the second row, magnified results of the palm (red box) demonstrate that our method produces smoother pressure transitions, more closely aligning with the ground truth (GT).
    }
    \label{fig:figure4}
\end{figure*}

\section{Experiments}
\label{sec:Experiments}

\subsection{Experimental Settings}

\noindent \textbf{Benchmarks.}
We evaluate our method on the EgoPressure dataset \cite{Egopressure}.
The dataset contains interactions from 21 participants, with each participant performing 64 interaction clips that have an average length of 420 frames. Images were captured by a system of one head-mounted egocentric camera and seven static RGB-D cameras.
The dataset provides pressure data captured from a touchpad (Sensel Morph \cite{SenselMorph}) and makes it available in a format that directly corresponds to the MANO \cite{mano} hand model's geometry. Figure~\ref{fig:figure2} (d) illustrates the process of generating ground truth UV-pressure maps. 

\noindent \textbf{Baselines.}
For comparison, we evaluate our method against several key baselines: PressureVision \cite{PressureVision}, PressureVision with HaMeR \cite{Egopressure}, PressureVision++ \cite{Pressurevision++}, and PressureFormer \cite{Egopressure}. 
The raw outputs from all baselines are image-based pressure maps, which we project onto a UV map using the pipeline proposed in EgoPressure.

\noindent \textbf{Evaluation Metrics.}
We evaluate our model using four metrics established in PressureVision \cite{PressureVision}. 
These include Contact IoU, which measures spatial accuracy via the IoU of binarized contact maps; 
Volumetric IoU, an extension that considers pressure magnitude by treating pressure maps as 3D volumes; 
Mean Absolute Error (MAE) in kPa between predicted and ground truth maps ; 
and Temporal Accuracy, the percentage of frames with a correctly predicted binary contact state. 
For all methods, the IoU-based metrics are computed directly on the UV-pressure maps.

\noindent \textbf{Implementation Details.}
Following the EgoPressure \cite{Egopressure} protocol, we use data from 15 participants for training and the remaining 6 for testing. We generate depth maps using Video Depth Anything \cite{VDA} and derive the remaining control signals from the EgoPressure annotations, with all frames resized to 256×256. The U-Net \cite{UNet} is initialized from pretrained SVD \cite{svd}, while the PoseNet and Vertex Encoder are trained from scratch. We train the model for 40k steps on 4 NVIDIA L20 48G GPUs with 16-frame sequences and a batch size of 2 per GPU, using a learning rate of 1e-5.
The target UV-pressure maps for the training process are denoised to remove artifacts and noise introduced by the physical sensor in the raw data.

\subsection{Comparison with Baselines}

\renewcommand{\arraystretch}{1.1} 
\begin{table}[tb]
\centering
    \newcommand{\Frst}[1]{\textbf{#1}}
    \newcommand{\Scnd}[1]{\underline{#1}}
    \newcommand{\gray}[1]{\textcolor{gray!60}{#1}}
    \caption{Quantitative comparison on the EgoPressure dataset. \textbf{Bold} text indicates the best result, while \underline{underlined} text indicates the second-best. ↑ denotes that higher values are better. 
    ``All Views" refers to models trained and tested on data from all eight cameras in the EgoPressure dataset, while the ``Ego View" means models used data exclusively from the head-mounted egocentric camera.}
    \label{tab:table1}
    \resizebox{0.48\textwidth}{!}
    {
    \LARGE
    \setlength{\tabcolsep}{3pt}
    \begin{tabular}{c | l | *{5}{c}}
        \toprule
        \textbf{\makecell[c]{Setting}} & \textbf{Models}  & \textbf{\makecell[c]{Contact\\IoU(\%)}} ↑ & \textbf{\makecell[c]{Vol.\\IoU(\%)}} ↑ & \makecell[c]{\textbf{MAE}\\(kPa)} ↓ & \textbf{\makecell[c]{Temporal\\Acc.(\%)}} ↑ \\
        \midrule
        \multirow{4}{*}{\makecell[c]{All\\Views}} 
        & PressureVision \cite{PressureVision} \small{ECCV'22}          & 21.53 & 16.41 & 44 & 90  \\
        & \cite{PressureVision} w. HaMeR \cite{hamer} \small{CVPR'25} & 24.10 & 17.36 & 49 & 92  \\ 
        & PressureVision++ \cite{Pressurevision++} \small{WACV'24}      & -     & -     & -  & -   \\
        & PressureFormer \cite{Egopressure} \small{CVPR'25}                & 33.12 & 24.54 & 71 & 89  \\
        \midrule
        \multirow{4}{*}{\makecell[c]{Ego\\View}} 
        & PressureVision \cite{PressureVision} \small{ECCV'22}          & 28.97 & 20.02 & 53 & \Scnd{92}  \\
        & \cite{PressureVision} w. HaMeR \cite{hamer} \small{CVPR'25} & 29.48 & 21.45 & 55 & \Scnd{92}  \\ 
        & PressureVision++ \cite{Pressurevision++} \small{WACV'24}      & \Scnd{32.25} & \Scnd{24.19} & \Scnd{48} & \Frst{94}  \\
        \rowcolor{gray!15}
        & EgoPressDiff \textbf{(Ours)}                                     & \Frst{39.53} & \Frst{32.61} & \Frst{43} & \Frst{94}  \\
        \bottomrule
    \end{tabular}
    }
    
\end{table}

\noindent \textbf{Quantitative Comparison.}
Table~\ref{tab:table1} presents the quantitative comparison results. As shown, EgoPressDiff achieves state-of-the-art performance. Notably, our method achieves a significant improvement in Volumetric IoU, outperforming the strongest baseline by over 34\%. Since the implementation for PressureFormer \cite{Egopressure} is not publicly available, we report its results on the ``All Views" setting directly from its original publication. For all other baselines, we present results on the ``Ego View" setting, which we obtained by retraining and evaluating their official public codebases on the egocentric view data.

\noindent \textbf{Qualitative Comparison.}
Figure~\ref{fig:figure4} presents a qualitative comparison across different methods.
The first row illustrates a scenario where the middle finger makes contact with the surface while the index finger does not, as confirmed by the side view. Baseline methods erroneously predict pressure on the non-contacting index finger. In contrast, our method, which incorporates depth, hand geometry and temporal information, yields the correct prediction.
As highlighted in the red box in the second row, our diffusion-based model generates a UV-pressure map with smoother pressure transitions, closely aligning with the ground truth. This is because baseline methods typically quantize pressure into a fixed number of classes (e.g., nine) and treat the problem as a pixel-wise classification task, resulting in coarse pressure gradients.
Overall, our method demonstrates superior performance in terms of both the accuracy of the contact regions and the fidelity of the pressure distribution.

\subsection{Ablation Study}
Our ablation study, presented in Table~\ref{tab:table2}, validates the contribution of each component in EgoPressDiff. We observe that removing any of the control signals (pose, depth, or vertices) degrades performance. Notably, the exclusion of depth input (Setting 2) causes the most significant drop across all metrics, especially in Temporal Accuracy, underscoring its critical role for inferring physical contact and maintaining temporal accuracy. Furthermore, naively fusing features without DC Spatial Layer (Setting 4) results in a more severe performance drop than removing the vertex input and its associated components (Vertex Encoder and DC Spatial Layer) entirely (Setting 3), which strongly validates our feature calibration strategy. Finally, removing the UV mask from the loss function (Setting 5) impairs spatial accuracy (IoUs), confirming that focusing the training on the valid hand region is crucial for high-fidelity reconstruction.

\renewcommand{\arraystretch}{1.1} 
\begin{table}[tb]
\centering
\caption{Ablation study for EgoPressDiff.}
    \resizebox{0.48\textwidth}{!}
    {
    \large
    \begin{tabular}{c | l | *{4}{c}}
        \toprule
        \textbf{ID} & \textbf{Settings}           & \textbf{\makecell[c]{Contact\\IoU (\%)}} ↑ & \textbf{\makecell[c]{Volumetric\\IoU (\%)}} ↑ & \makecell[c]{\textbf{MAE}\\(kPa)} ↓ & \textbf{\makecell[c]{Temporal\\Acc. (\%)}} ↑ \\
        \midrule
        1 & \textit{w/o} Pose Input         & 38.45     & 31.25     & 48    & 94 \\
        2 & \textit{w/o} Depth Input        & 28.57     & 19.83     & 62    & 82 \\
        3 & \textit{w/o} Vertices Input     & 36.21     & 29.70     & 52    & 90 \\
        4 & \textit{w/o} DC Spatial Layer   & 34.14     & 27.98     & 51    & 87 \\
        5 & \textit{w/o} UV Mask            & 32.56     & 23.17     & 55    & 91 \\
        \rowcolor{gray!15}
        - & EgoPressDiff \textbf{(Ours)}    & 39.53     & 32.61     & 43    & 94 \\
        \bottomrule
    \end{tabular}
    }
    \label{tab:table2}
\end{table}

\section{Conclusion}
In this work, we reframed egocentric hand-pressure estimation as continuous video generation and introduced EgoPressDiff, a multi-modal video diffusion model that generates UV-pressure maps from visual input. 
By conditioning on complementary signals and aligning their statistics through the proposed modules, our method produces plausible UV-pressure maps directly on the hand mesh.
Taken together, these design choices yield state-of-the-art results on the EgoPressure ego-view setting, with consistent gains across all metrics.
However, our model has some limitations. It is trained on simple hand postures and contact patterns, limiting generalization to complex daily activities. Future work includes building a more diverse hand-pressure dataset for daily activities and complex grasps, and extending EgoPressDiff with stronger priors and adaptation mechanisms to handle multi-contact scenarios.

\clearpage

\section{Acknowledgements}
This work was supported by the National Natural Science Foundation of China (U23B2030, Nos. 62311530100 and 62171251) and the Special Foundations for the Development of Strategic Emerging Industries of Shenzhen (No. KJZD20231023094700001).

\bibliographystyle{IEEEbib}
\bibliography{strings,refs}

\end{document}